\pdfoutput=1
\documentclass[a4paper,twoside]{article}
\usepackage{epsfig}
\usepackage{epstopdf}
\usepackage{subcaption}
\usepackage{calc}
\usepackage{amsmath,amsfonts}
\usepackage{amsmath}
\usepackage{multicol}
\usepackage{pslatex}
\usepackage{apalike}
\usepackage[algo2e]{algorithm2e}
\usepackage[bottom]{footmisc}
\usepackage{placeins}
\usepackage{bm,amssymb}
\usepackage{booktabs} 
\usepackage{xcolor}
\usepackage{tabularx}
\usepackage{algorithm}
\usepackage{algorithmic}
\usepackage{float}
\usepackage{SCITEPRESS}     

\newcommand{\mat}[1] {\mathbf{#1}}

\newcommand{\tns}[1] {\bm{\mathcal{#1}}}

\begin{document}

\title{Multidimensional Compressed Sensing for Spectral Light Field  Imaging\vspace{-10pt}}
\author{\authorname{Wen Cao, 
Ehsan Miandjiand and Jonas Unger}
Media and Information Technology,Department of Science and Technology, Linköping University, SE-601 74 Norrköping, Sweden\\
\email{\{f\_wen cao, s\_ehsan.miandjir\}@liu.se, t\_jonas.unger@liu.se}
}

\keywords{Spectral light field, Compressive sensing}

\abstract{This paper considers a compressive multi-spectral light field camera model that utilizes a one-hot spectral-coded mask and a microlens array to capture spatial, angular, and spectral information using a single monochrome sensor. We propose a model that employs compressed sensing techniques to reconstruct the complete multi-spectral light field from undersampled measurements. Unlike previous work where a light field is vectorized to a 1D signal, our method employs a 5D basis and a novel 5D measurement model, hence, matching the intrinsic dimensionality of multispectral light fields. We mathematically and empirically show the equivalence of 5D and 1D sensing models, and most importantly that the 5D framework achieves orders of magnitude faster reconstruction while requiring a small fraction of the memory. Moreover, our new multidimensional sensing model opens new research directions for designing efficient visual data acquisition algorithms and hardware.}
\onecolumn \maketitle \normalsize
\setcounter{footnote}{0} \vfill
\section{\uppercase{Introduction}}
\label{sec:introduction}

\begin{figure*}[ht]
 \includegraphics[width=\textwidth]{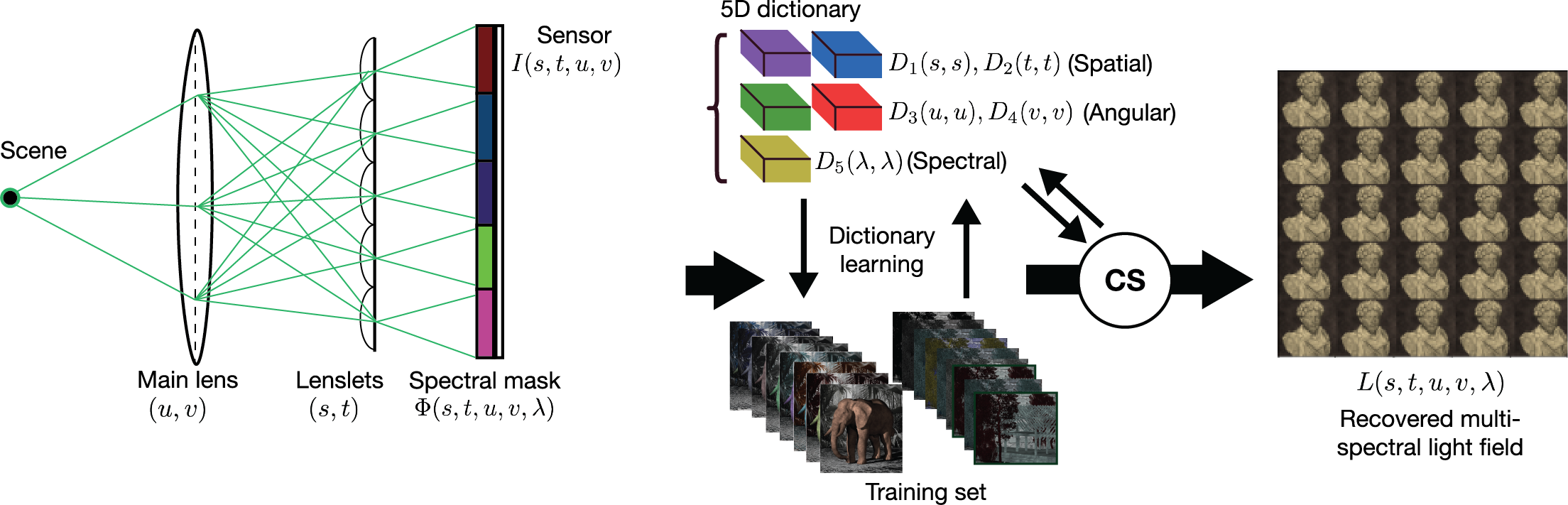}\vspace{-0 mm}
\caption{Illustrates the proposed compressed sensing framework for multi-spectral light field capture and reconstruction. Compressed light fields are captured using a lenslet array placed in the optical path and a one-hot spectral CFA mask placed on the sensor. Our proposed $nD$ compressed sensing formulation improves the reconstruction time by orders of magnitude as compared to the commonly used 1D compressed sensing techniques without any quality degradation.}
  \label{Fig_Dia}
   \end{figure*}
Computational cameras for light field capture, \cite{levoy_light_1996,gortler_lumigraph_1996}, post capture editing and scene analysis have become increasingly popular and found applications ranging from photography and computer vision to capture of neural radiance fields. Light field imaging aims to obtain multidimensional optical information including spatial, angular, spectral, and temporal sampling of the scene. Inherent to most light field capture systems is the trade-off between the complexity and cost of the acquisition system and the output image quality in terms of spatial and angular resolution. On one hand camera arrays, \cite{10.1145/1073204.1073259}, offer high resolution but comes with high cost and complex bulky mechanical setups, while systems based on microlens, or lenslet, arrays placed in the optical path, \cite{ng:hal-02551481}, sacrifices resolution for the benefit of light weight systems and lower costs. 

A key goal in the development of next generation light field imaging systems is to enable high quality multispectral measurements of complex scenes based on a minimum amount of input measurements. Focusing on cameras with lenslet arrays, such optical systems capture 2D projections of the full 5D (spatial, angular, spectral) light field image data. Reconstruction of the high dimensional 5D data from the 2D measurements is challenging, due to the dimensionality gap and high compression ratio resulting from undersampling of, e.g., the spectral domain. Compressed sensing (CS) has emerged as a popular approach for light field reconstruction, however, current systems are either not designed for multispectral reconstruction, \cite{marwah_compressive_2013,miandji_unified_2019}, and/or fundamentally rely on 1D CS reconstruction, \cite{marquez_compressive_2020}, disregarding the original signal dimensionality. Since a light field is fundamentally a 5D object, a vectorized 1D representation of such data will prohibit the exploitation of data coherence along each dimension, which has been observed in a number of previous works on sparse representation of multidimensional data \cite{miandji_unified_2019}. Another problem with 1D reconstruction is that it inherently leads to prohibitively large storage costs for the dictionary and sensing matrices as well as long reconstruction times. 

In this paper, we formulate the light field capture and reconstruction as a multidimensional, $nD$, compressed sensing problem. As a first step, see Fig.~\ref{Fig_Dia}, we learn a 5D dictionary ensemble \cite{miandji_unified_2019} from the spatial (2D), angular (2D), and spectral (1D) domains of a light field training set. Using a novel 5D sensing model based on a one-hot sampling pattern (implemented as a multispectral color filter array (CFA) on the sensor), we obtain measurements of a light field in the test set. The 5D sensing model is composed of 5 measurement matrices, each corresponding to one dimension of the light field. By randomizing the one-hot measurement for each measurement matrix, we promote the incoherence of the measurement matrices with respect to the 5D dictionary. 
The one-hot spectral sampling mask allows us to design a spectral light field camera with a monochrome sensor. Another advantage of the one-hot mask is cost-effectiveness, since it has a similar manufacturing complexity compared to a Color Filter Array (CFA), which is commonly used in consumer-level digital cameras. Finally, we extend the Smoothed-$\ell_0$ (SL0) method for 2D signals \cite{2dsl0} to 5D signals, enabling fast reconstructions of the light field from the measurements without the need for manipulating the dimensionality of the light field.

The main contributions are: 

\begin{itemize}
\item A novel $nD$ formulation for a single sensor compressive spectral light field camera design, where a one-hot sensing model together with a learned multidimensional sparse representation is utilized. 
\item An $nD$ recovery method that is more than two orders of magnitude faster than the widely used 1D formulation and recovery.
\item We show, both theoretically and experimentally, that the $n$D formulation is far superior to the $1$D variant both in terms of memory and speed.
\end{itemize}

Our sensing model derivation shows that the $n$D sensing is mathematically equivalent to 1D sensing with the same number of samples. Experimental results confirm such equivalence in terms of reconstruction quality. Most importantly, the evaluations show that the proposed novel formulation produces high quality results orders of magnitude faster compared to methods based on 1D compressed sensing. Our multidimensional sensing model opens up flexible sensing mask designs where each dimension can be treated individually. 

\section{\uppercase{Related work}}
\label{sec1}
Common approaches for light field imaging include the use of coded apertures~\cite{10.1145/1360612.1360654,10.5555/1819298.1819383} and lenslet arrays \cite{ng:hal-02551481}, and more recently combinations of the two~\cite{marquez_compressive_2020}. Combining compressive sensing with spectrometers, coded aperture snapshot spectral imagers are also known as compressive spectral imagers (CASSI). In general, a CASSI system consists of a coded mask, prisms, and an imaging sensor. Examples of capture systems are for example described by~\cite{hua_ultra-compact_2022}, and~\cite{xiong_snapshot_2017}. Based on the detector measurements and the coded mask information the final spectral images or light fields are then typically reconstructed using compressed sensing methods~\cite{arce_compressive_2014,marquez_compressive_2020,yuan_snapshot_2021}. Recently, deep learning based methods have shown promising results for the application of multispectral image reconstruction, for an overview, see the survey by~\cite{huang_spectral_2022}. Related to our approach~\cite{schambach_spectral_2021} proposed a multi-task deep learning method to estimate the center view and depth from the coded measurements. The approach yields good results, but is not directly comparable to our approach since the aim here is to recover the full 5D light field and not a depth map. 

Compared to previous work, our $nD$ formulation presents several benefits such as flexible design of the 5D light field sensing masks as well as orders of magnitude speed up without any quality degradation as compared to the 1D sensing used in previous work. One of the key insights, and contributions, enabling the $nD$ CS framework is the sensing mask formulation illustrated in Fig.~\ref{fig_product}, where the Kronecker product of a set of small design matrices lead to the one-hot CFA mask used for $nD$ compressed sampling. This as well as our general formulation are described in detail in the next section.

 \section{Multidimensional Compressive Light Fields}

Our capture system, see Fig.~\ref{Fig_Dia}, records a set of angular views as produced by the lenslet array, where a one-hot spectral coded mask at the resolution of the lenslet is placed on the sensor. In the design, we assume that there are $n$ different types of filters with different spectral characteristics available, and that the goal is to formulate a one-hot coded mask such that each measurement, or pixel on the sensor, integrates the incoming light rays modulated by one of the $n$ different filter types. This is motivated by the fact that we want to enable capture of a well defined set of spectral bands, typically defined by the imaging application. 

\subsection{The measurement model}\label{sec:measurement}

\begin{figure}[ht]
  \includegraphics[width=\linewidth]{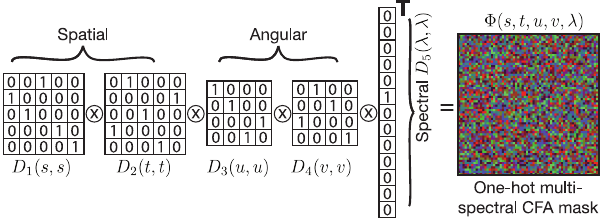}
   \caption{Illustration of Kronecker-based coded attenuation masks.}
     \label{fig_product}
     \vspace{-0mm}
\end{figure}

Our camera design measures the spatial and angular information, while the spectral information, due to the one-hot mask, is compressed to a single value. Let $L(s,t,u,v,\lambda)$ represent the light field function following the commonly used two-plane parameterization, where $(s,t)$ and $(u,v)$ denotes the spatial and angular dimensions, and $\lambda$ the spectral dimension. Moreover, let $I(s,t,u,v)$ be the compressed measurements recorded by the sensor. We formulate the multispectral compressive imaging pipeline as follows
\begin{equation}
\label{eq:imaging}
I(s,t,\mu,\nu) =\int_{\lambda} \Phi(s, t, u, v, \lambda) L(s,t, u, v,\lambda)\text{d} \lambda,
\end{equation}
where $\Phi(s, t, u, v, \lambda)$ is the 5D sensing operator, as shown in Fig.~\ref{fig_product}. Specifically, due to our design of the one-hot binary mask, we have
\begin{equation}
\label{eq:onehot}
\sum_{\lambda=1}^{n} \Phi[s, t,  u, v, \lambda]=1 ,
\end{equation}
making sure that only one spectral channel is sampled for every pixel on the monochrome sensor.

According to \eqref{eq:imaging}, our goal is to recover the full multispectral light field $L$ from its measurements $I$ and with the knowledge of the measurement operator $\Phi$ constructed from the one-hot mask. Indeed, since \eqref{eq:imaging} is a linear measurement model, we can rewrite it as $I=\Phi L$, where $L$ is the light field arranged in a vector and $\Phi$ is the \emph{measurement matrix}. In this case, recovering $L$ amounts to solving an underdetermined system of linear equations, which is a topic addressed by compressed sensing~\cite{candes_introduction_2008}. 

\subsection{Compressed sensing}\label{sec:cs}

The underdetermined system of linear equations $I=\Phi L$ can be solved by regularizing the problem based on sparsity of the light field in a suitable basis or \emph{dictionary} $D$ as follows
\begin{equation}
\vspace{-3pt}
 \mathop{argmin}\limits_{\alpha}{\|\alpha\|_0} \quad  \textrm{s.t.} \quad ||{I}-\Phi D \alpha||_2\le\epsilon,
 \label{eq:l0-1D}
 \vspace{-3pt}
\end{equation}
where $\epsilon$ is a user-defined threshold for the data fidelity and $\|.\|_0$ denotes the $\ell_0$ pseudo-norm. The dictionary $D$ can be obtained from analytical basis function, e.g., Discrete Cosine Transform (DCT), or by training-based algorithms, e.g., K-SVD \cite{aharon_k-svd_2006}. Equation \eqref{eq:l0-1D} can be solved by several algorithms and in this paper we use Smoothed-$\ell_0$ (SL0)~\cite{mohimani_sparse_2010}. We note that \eqref{eq:l0-1D} solves a 1D compressed sensing problem, and therefore utilizes a 1D dictionary. As mentioned, our goal is to solve \eqref{eq:l0-1D} using multidimensional dictionaries and a multidimensional sensing model. 

\subsection{Learning-based multidimensional dictionary}\label{sec:amde}

To this end, we utilize the Aggregated Multidimensional Dictionary Ensemble (AMDE) \cite{miandji_unified_2019} to train an ensemble of 5D orthogonal dictionaries to efficiently represent spatial, angular, and spectral dimensions of a light field. In particular, we have $D_1\in \mathbb{R}^{s \times s},D_2\in \mathbb{R}^{t \times t},D_3\in \mathbb{R}^{u \times u},D_4\in \mathbb{R}^{v \times v},D_5\in \mathbb{R}^{\lambda \times \lambda}$. The 1D dictionary, used in \eqref{eq:l0-1D}, can be obtained from a 5D dictionary using the Kronecker product \cite{duarte_kronecker_2012} as follows
\begin{equation}\label{eq:kron}
D = D_5\otimes D_4\otimes D_3\otimes D_2\otimes D_1,
\end{equation}
where $D \in \mathbb{R}^{stuv\lambda\times stuv\lambda}$. Indeed, the storage cost of $D$ is prohibitively large compared to the per-dimension dictionaries $D_i$. The memory requirement ratio between a 1D and a 5D dictionary is $\frac{(stuv\lambda)^2}{s^2+t^2+u^2+v^2+\lambda^2}$. This further motivates a multidimensional dictionary and sensing model.

\subsection{Multidimensional measurement model}\label{sec:nd-samp}

In multidimensional sensing, a requirement is to introduce a separate sensing operator for each dimension of the signal. Since a light field is 5D, we need to design 5 sensing operators to obtain measurements from the spatial, angular, and spectral domains. Let $\Phi_j$, $j=1,\dots,5$ be the set of sensing matrices (or operators) for a 5D light field. Utilizing the n-mode product between a tensor and a matrix \cite{tensor-kolda}, we can formulate the multidimensional measurement model as follows
\begin{equation}\label{eq:nd-sensing}
\vspace{-0pt}
\tns{I} = \tns{L} \times_1\Phi_1 \times_2\Phi_2 \times_3 \Phi_3 \times_4 \Phi_4 \times_5 \Phi_5
\vspace{-0pt}
\end{equation}
where $\tns{L}\in\mathbb{R}^{s\times t\times u\times v\times\lambda}$ is the light field tensor and $\tns{I}\in\mathbb{R}^{s\times t\times u\times v}$ is the compressed measurements on the sensor. The choice of sensing matrices $\Phi_j$ is of utmost importance in any compressive acquisition setup. Due to our camera design in Fig.~\ref{Fig_Dia}, i.e. the use of a lenslet array, we do not perform compression on spatial and angular domains. As a result, $\Phi_j$, $j=1,\dots,4$, are identity matrices. 
However, we would like to maximize the incoherence between the dictionary and the sensing matrices \cite{candes_introduction_2008}. Therefore, as illustrated in Fig.~\ref{fig_product}, we perform a random row-shuffling of the identity matrices to form $\Phi_j$, $j=1,\dots,4$. 
On the other hand, since we perform spectral compression using the one-hot mask, $\Phi_5\in\mathbb{R}^{1\times n}$ contains only one nonzero value, where $n$ is the total number of spectral bands. In our experiments, we used data sets where $n=13$. 

The multidimensional measurement model in \eqref{eq:nd-sensing} can be converted to a 1D measurement model as follows
\begin{equation}\label{eq:kron-sense}
    I = \underbrace{(\Phi_5\otimes\Phi_4\otimes\Phi_3\otimes\Phi_2\otimes\Phi_1)}_{\Phi} L,
    \vspace{-3pt}
\end{equation}
where L and I are vectorized forms of the tensors $\tns{L}$ and $\tns{I}$, respectively. An illustration of the Kronecker product of sensing matrices $\Phi_j$ is shown in Fig.~\ref{fig_product}. Note that the key benefit of the nD sensing model is that it requires several orders of magnitude less memory compared to the Kronecker 1D sensing model in~\eqref{eq:kron-sense}. Recall from Section~\ref{sec:amde} that our 5D AMDE dictionary also exhibits such benefits. The nD approach also comes with the benefit that the sensing mask can be designed using only five small matrices.

The pseudocode in Algorithm~\ref{alg:sl0gpu} illustrates nD SL0 reconstruction given the measurements $\tns{I}$ and the matrices $\mat{A}^{(i)}=\Phi_iD_i$ as presented in Section 3.5.

             
              

\begin{algorithm}
\caption{Multidimensional SL0 algorithm}
\begin{algorithmic}[1]
\label{alg:sl0gpu}
\REQUIRE Input tensor $\tns{I}$, input matrices $\mat{A}^{(1 \ldots n)}$, limit parameter $\sigma_{min}$, decreasing factor $\sigma_{f}$, iteration $L$, step $\mu$

\ENSURE $\tns{S}$ as sparse solution of $\tns{S} \times_{1 \ldots n} \mat{A}^{(1 \ldots n)} = \tns{I}$ 

\STATE $\tns{S} \leftarrow \tns{I} \times_{1 \ldots n} \mat{A}^{(1 \ldots n)\dagger} $ \quad \quad \COMMENT{initial estimate, $\dagger$ denotes pseudo-inverse} 

\STATE $\sigma \leftarrow$ largest absolute element of $\tns{I}$

\WHILE{$\sigma > \sigma_{min}$}
\FOR{$k = 1 \ldots L$}

\STATE $\Delta\tns{S} = -\frac{\tns{S}}{\sigma^2} \circ exp\Big(-\frac{\tns{S} \circ \tns{S}}{2\sigma^2}\Big) $

\STATE $\hat{\tns{S}} \leftarrow \mu \cdot \Delta \tns{S}$ \quad \quad \COMMENT{Steepest ascent step}

\STATE $\hat{\tns{S}} \leftarrow \hat{\tns{S}} -  (\hat{\tns{S}} \times_{1 \ldots n} \mat{A}^{(1 \ldots n)} - \tns{I}) \times_{1 \ldots n} \mat{A}^{(1 \ldots n)\dagger}$

\ENDFOR

\STATE $\tns{S} \leftarrow \hat{\tns{S}}$

\STATE $\sigma = \sigma_f \cdot \sigma$

\ENDWHILE
\end{algorithmic}
\end{algorithm}

\subsection{Reconstruction}

Given the AMDE dictionary in Section~\ref{sec:amde} and the multidimensional measurement model in Section~\ref{sec:nd-samp}, we formulate the 5D light field recovery algorithm from its measurements as follows 
\begin{multline}\label{eq:5d-recon}
\vspace{-5pt}
\mathop{argmin}\limits_{\tns{S}}\|\tns{S}\|_0 \quad  \textrm{s.t.} \quad \|\tns{I}-\tns{S}\times_1\Phi_1D_1\times_2\Phi_2D_2\times_3 \\ \Phi_3D_3\times_4\Phi_4D_4\times_5\Phi_5D_5 \|_2\le\epsilon,
\vspace{-5pt}
\end{multline}
where $\tns{S}\in\mathbb{R}^{s\times t\times u\times v\times \lambda}$ is a sparse coefficient vector. Once the coefficients are obtained, the full light field is computed as $\hat{\tns{L}}=\tns{S}\times_1D_1\times_2D_2\times_3D_3\times_4D_4\times_5D_5$. To solve \eqref{eq:5d-recon}, we extend the 2D SL0 algorithm \cite{2dsl0} for higher dimensional signals. A pseudo-code of the 5D SL0 algorithm is provided in Algorithm \ref{alg:sl0gpu}. 

\section{Evaluation and results}
\begin{table*}[]%
\caption{Comparing 5D DCT, 1D AMDE and 5D AMDE using five test spectral light fields under ONE snapshot. Proposed Method in blue. Best values in bold.}
\small
\label{tab_results}
\begin{center}
\begin{tabular*}{\textwidth}{@{\extracolsep{\fill}}l|lll|lll|lll}
  \toprule
 \textbf {Test} &\textbf {Methods}&PSNR /dB& &SSIM&&  &SA/{($^{\circ}$)}&&\\
  \cmidrule(lr){1-10}
  \textbf { Scenes}&5D&1D&  \textcolor{blue}{5D}&5D&1D& \textcolor{blue}{5D}&5D&1D&  \textcolor{blue}{5D}\\ 
  &DCT&AMDE&  \textcolor{blue}{AMDE}&DCT&AMDE& \textcolor{blue}{AMDE}&DCT&AMDE&  \textcolor{blue}{AMDE}\\
  \midrule\midrule[.1em]
Bust  & 25.388 &31.944 &31.944 &0.741&0.812&0.812&22.6658&10.853&10.853\\ 
 Cabin  &15.798 & 20.275&20.2754&0.302&0.636&0.636&53.767&26.977&26.977\\
  Circles &15.459 &18.494  &18.494&0.313&0.530&0.530&35.216
&29.588&29.588\\
  Dots & 15.454&23.485 &23.485&0.247&0.738&0.738& 24.661& 10.668& 10.668\\
  Elephants &18.108&24.880&24.880&0.5761&0.743&0.7435&23.161&17.356&17.356\\
  \cmidrule(lr){1-10}
 \textbf { Average} &14.631& \textbf{22.398} & \textbf{22.398}&0.324&\textbf{0.627}&\textbf{0.627}&35.54&\textbf{24.804}&\textbf{24.804}\\ 
  \bottomrule
\end{tabular*}
\end{center}
\bigskip\centering
\vspace{-5pt}
\end{table*}%

We evaluate our multispectral light field camera design and CS framework using the multispectral data set published by~\cite{schambach_multispectral_2020}. As training set we randomly choose 60 out of the 500 random scenes, and as test set we use the five handcrafted scenes. The [$512\times 512 \times 5 \times 5 \times 13$] light fields exhibit a spatial resolution of $512\times512$ pixels and an angular resolution of $5\times5$ views sampled in 13 spectral bands in steps of 25 nm from 400 nm to 700 nm. For all experiments we use a patch size of $5 \times 5 \times 4 \times 4 \times 13$, where the one-hot encoded mask illustrated in Fig.~\ref{fig_product} samples the spectral domain using a single sample per pixel, leading to a compression ratio of $1/13$ corresponding to $7.69$\% of the original samples. We also report our results with multiple shots, i.e. when multiple shots are taken by the camera system in Fig. \ref{Fig_Dia}, where the mask pattern is changed for each shot.

\begin{table}[ht]
  \caption{ Average Reconstruction Time of 5D DCT, 1D AMDE and 5D AMDE per spectral light  field using ONE snapshot}
    \centering
    \begin{tabular}{@{} p{2.0cm} *{2}{>{\ttfamily}l} @{}}
    \toprule[.1em]
    Methods & \normalfont Reconstruction Time \\ 
    \midrule\midrule[.1em]
     {5D DCT}& \normalfont{ 40.3 seconds} \\ 
      \cmidrule(lr){1-2}
      {1D AMDE} & \normalfont{2.4 hour}  \\ 
      \cmidrule(lr){1-2}
     {5D AMDE}& \normalfont{79.5 seconds} \\  
    \bottomrule[.1em]
    \end{tabular}
    \label{tab2}
    \end{table} 

Table~\ref{tab_results} compares 5D CS using the AMDE basis described above to 1D CS using AMDE in terms of peak signal-to-noise ratio (PSNR),
structural similarity index measure (SSIM), and spectral
angle (SA). We also include 5D DCT as representative for commonly used analytical bases in compressed sensing. The results show that our novel $nD$ formulation and the resulting 5D multidimensional sensing mask perform as expected with PSNR, SSIM, and SA on par with the 1D approach. The important difference, however, is that the $nD$ formulation is orders of magnitude faster, specifically 106 times faster, running on a machine with 16 CPU cores operating at 4.5GHz. The average reconstruction time comparison of the 5D CS and 1D CS is provided in Table~\ref{tab2}.
More importantly, our 5D sensing and reconstruction require only a fraction of the memory, specifically $\frac{1}{107729} = 9.2825e-06$.

\begin{figure*}[ht]
    \setlength{\tabcolsep}{0.01cm}
    \begin{tabularx}{0.90\linewidth}{ccccc}
        \small\textbf{Ground Truth} & \small\textbf{1 Snapshot}  & \small\textbf{3 Snapshots}  & \small\textbf{5 Snapshots} &\small\textbf{7 Snapshots}\\ 
        
        \small\textbf{\rotatebox{90}{5D DCT}}
        \includegraphics[height=0.95in,width=0.185\linewidth]{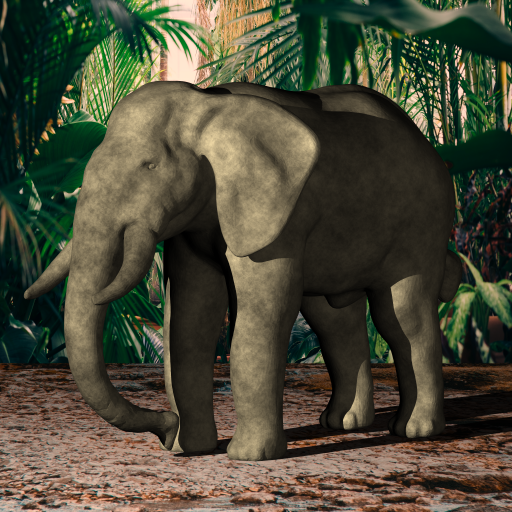} &
        \includegraphics[height=0.95in,width=0.185\linewidth]{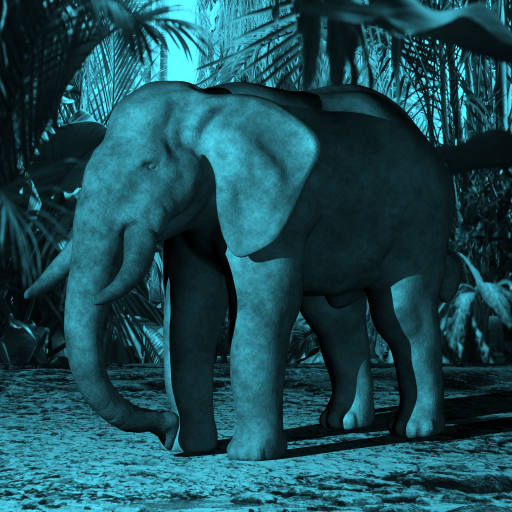} &
        \includegraphics[height=0.95in,width=0.185\linewidth]{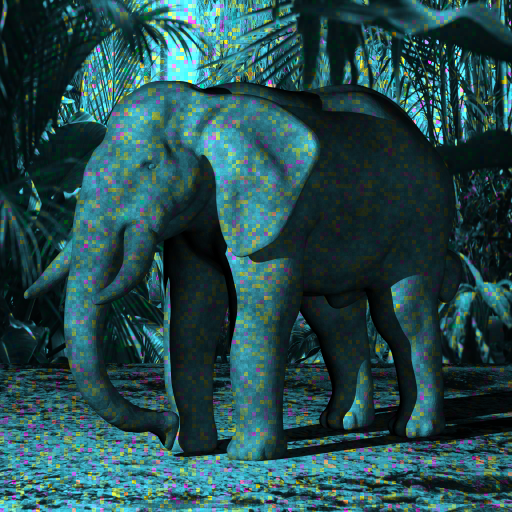} &
        \includegraphics[height=0.95in,width=0.185\linewidth]{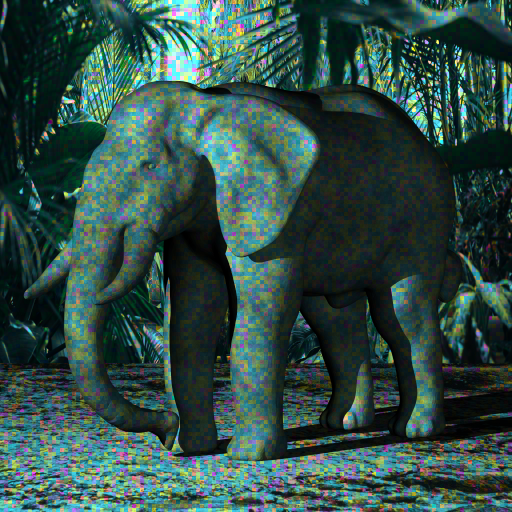} &\includegraphics[height=0.95in,width=0.185\linewidth]{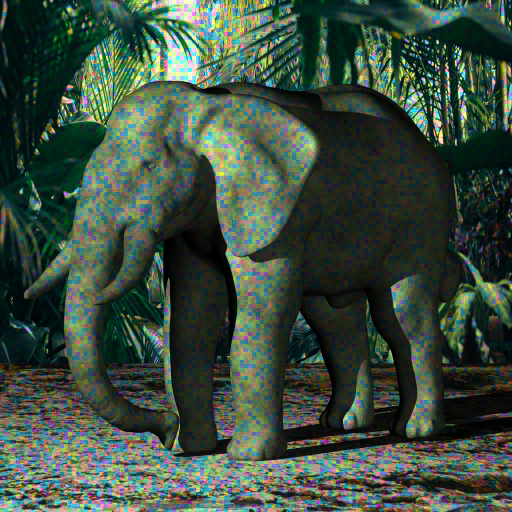}\\
        PSNR&18.1089 &18.3892&19.4972&21.2688\\
        \small\textbf{\rotatebox{90}{5D AMDE}}
        \includegraphics[height=0.95in,width=0.185\linewidth]{Elephant.png}  &
        \includegraphics[height=0.95in,width=0.185\linewidth]{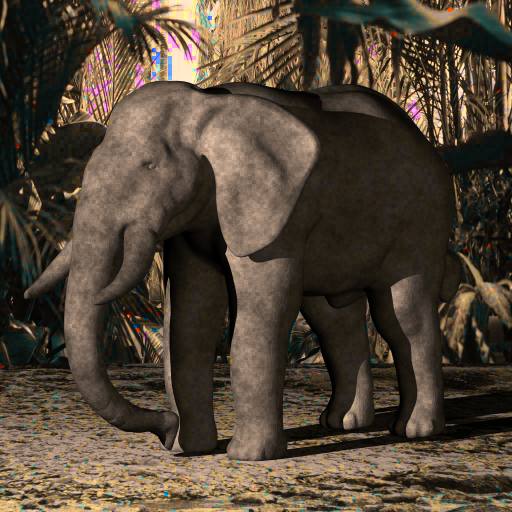} &
        \includegraphics[height=0.95in,width=0.185\linewidth]{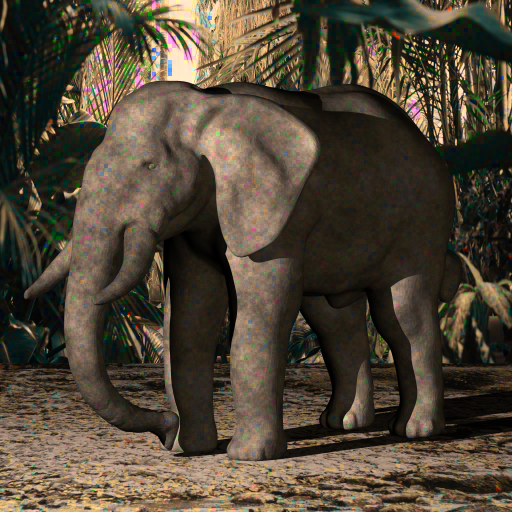}& \includegraphics[height=0.95in,width=0.185\linewidth]{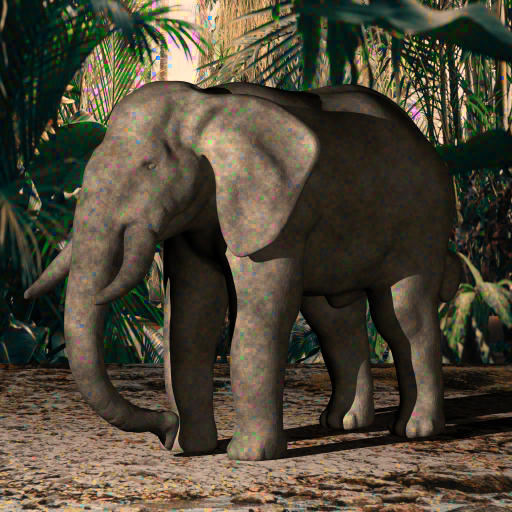} &\includegraphics[height=0.95in,width=0.185\linewidth]{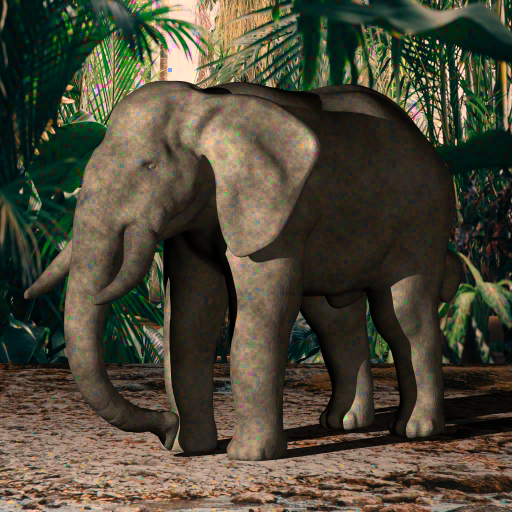}\\
       PSNR &24.8808&25.0373&26.4341&\textbf{28.7968}\\      
      \end{tabularx}
    \caption{5D CS Performance comparison (5D DCT, 5D AMDE) of reconstruction results using different number of snapshots of Elephant scene. Captions are identical to Fig.~\ref{figss}.}
    \label{fig33}
\end{figure*}

Generally, an imaging system can take more snapshots to get more measurements. This can yield better results for compressive sensing methods ~\cite{yuan_snapshot_2021}. 
As all previous evaluations were conducted under a single snapshot condition, we can significantly enhance the reconstruction quality by employing additional snapshots as shown in Fig.~\ref{fig33}.  Specifically, when utilizing five snapshots for recovering the elephant scene with the 5D AMDE basis, the recovered spectral light field exhibits accurate colors. Although the 5D DCT basis also shows improved reconstruction with an increased number of snapshots, it fails in accurately reproducing colors.

\begin{figure*}[ht]
    \setlength{\tabcolsep}{0.01cm}
    \begin{tabularx}{0.90\linewidth}{cccc}
        \small\textbf{Ground Truth} & \small\textbf{5D DCT}  & \small\textbf{1D AMDE}  & \small\textbf{5D AMDE} \\ 
        \small\textbf{\rotatebox{90}{Bust}}
        \includegraphics[height=1.2in,width=0.230\linewidth]{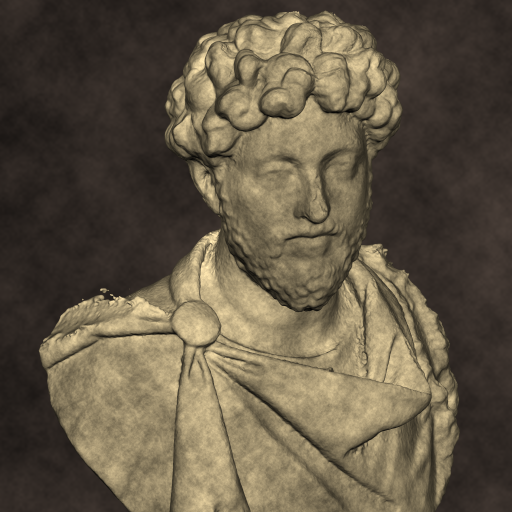} &
        \includegraphics[height=1.2in,width=0.230\linewidth]{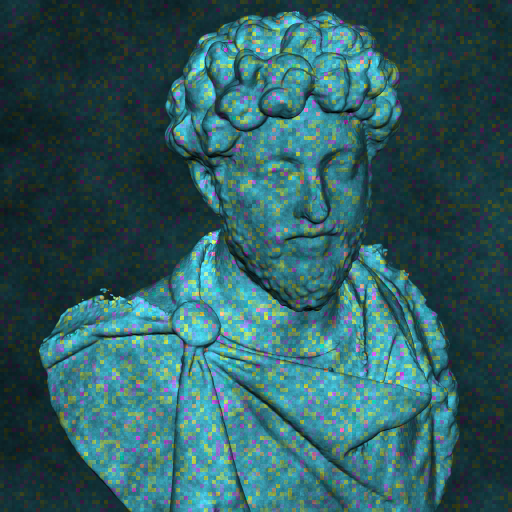} &
        \includegraphics[height=1.2in,width=0.230\linewidth]{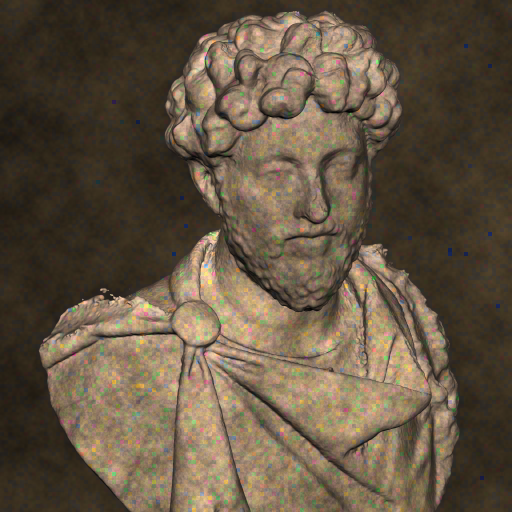} &
        \includegraphics[height=1.2in,width=0.230\linewidth]{Bust_3shots.png} \\
          PSNR&25.4520&32.1021&32.1021\\ 
        
        \small\textbf{\rotatebox{90}{Cabin}}
        \includegraphics[height=1.2in,width=0.230\linewidth]{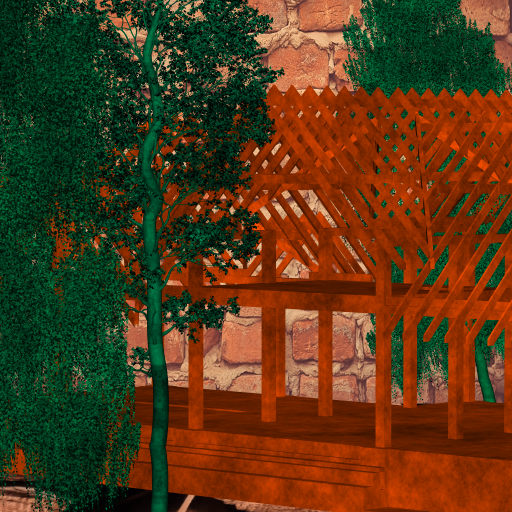} &
        \includegraphics[height=1.2in,width=0.230\linewidth]{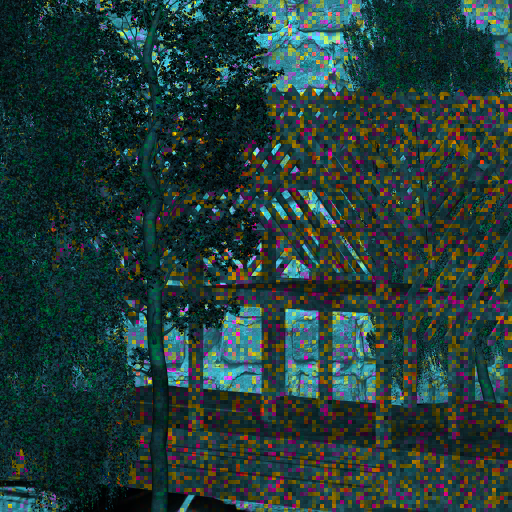} &
        \includegraphics[height=1.2in,width=0.230\linewidth]{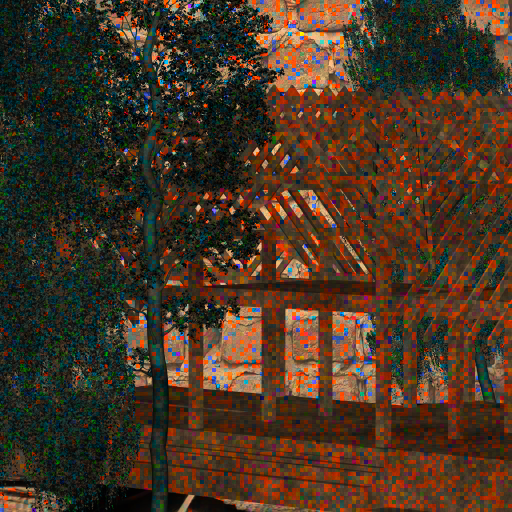} &
        \includegraphics[height=1.2in,width=0.230\linewidth]{Cabin_3shots.png} \\
        PSNR&16.1041&21.5743&21.5743\\ 
        
        \small\textbf{\rotatebox{90}{Circles}}
        \includegraphics[height=1.2in,width=0.230\linewidth]{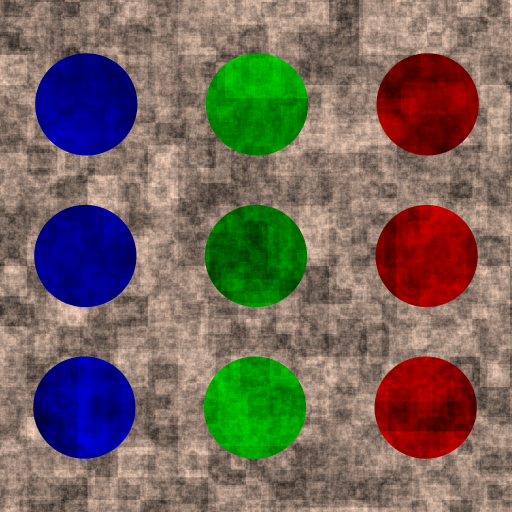} &
        \includegraphics[height=1.2in,width=0.230\linewidth]{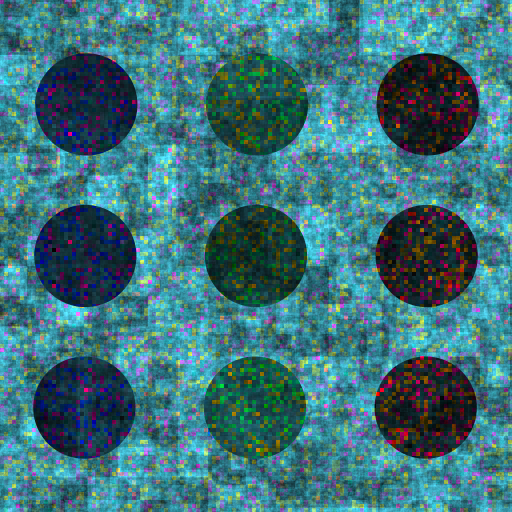} &
        \includegraphics[height=1.2in,width=0.230\linewidth]{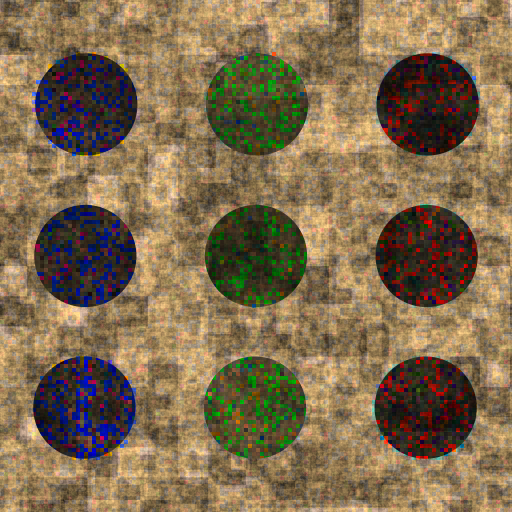} &
        \includegraphics[height=1.2in,width=0.230\linewidth]{Circle_3shots.png} \\
          PSNR&15.6698&18.9974&18.9974\\ 

        \small\textbf{\rotatebox{90}{Dots}}
        \includegraphics[height=1.2in,width=0.230\linewidth]{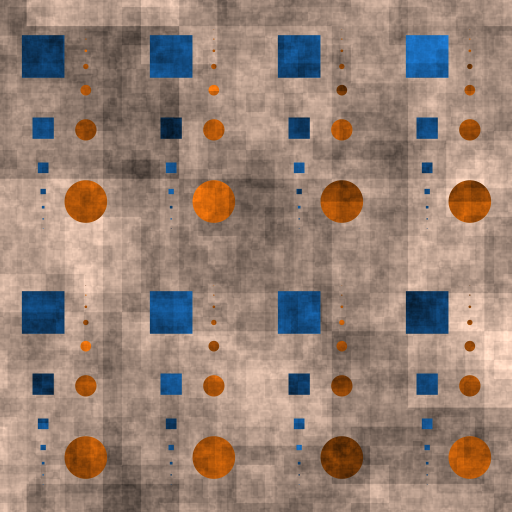} &
        \includegraphics[height=1.2in,width=0.230\linewidth]{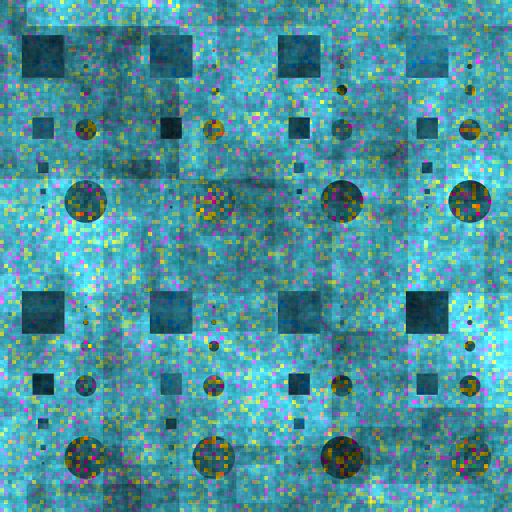} &
        \includegraphics[height=1.2in,width=0.230\linewidth]{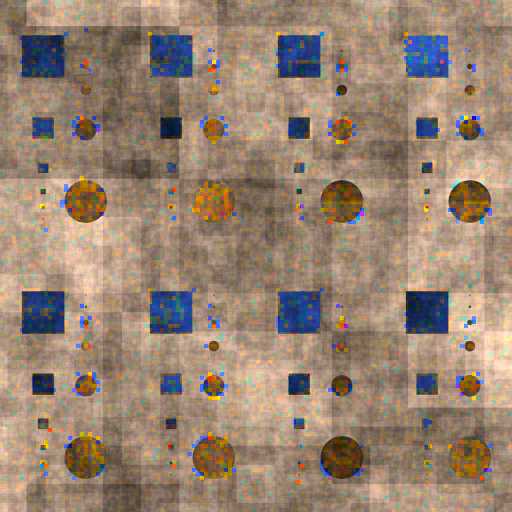} &
        \includegraphics[height=1.2in,width=0.230\linewidth]{dots_3shots.png} \\
          PSNR&16.1914&24.7977&24.7977\\  

        \small\textbf{\rotatebox{90}{Elephant}}
        \includegraphics[height=1.2in,width=0.230\linewidth]{Elephant.png} &
        \includegraphics[height=1.2in,width=0.230\linewidth]{Elep_3shots_dct.png} &
        \includegraphics[height=1.2in,width=0.230\linewidth]{Elep_3shots.png} &
        \includegraphics[height=1.2in,width=0.230\linewidth]{Elep_3shots.png} \\
     PSNR&18.3892&25.0373&25.0373\\
        
    \end{tabularx}
    \caption{Performance comparison of Reconstruction Results of 5D DCT, 1D AMDE and 5D AMDE of five test spectral light fields using three snapshots. Evaluation metric PSNR in dB. The multi-spectral channels are converted to RGB according to CIE 1913 and CIE D65. The images are chosen as the angular image [2 3] out of the $5\times5$ reconstructed views. }
    \label{figss}
\end{figure*}

Furthermore, Fig.~\ref{figss} presents a  comparative visualization of the reconstruction results of the spectral light field for all scenes in Table~\ref{tab_results} using three snapshots. As expected, 5D CS using the AMDE basis achieves visually identical results when compared to 1D AMDE. And 5D AMDE outperforms the 5D CS with the DCT basis as 5D DCT cannot faithfully recover spectral information.

\section{Conclusions}

This paper introduced a sensing and reconstruction method for fast multispectral light field acquisition. We believe that our novel $n$D formulation, and in particular the one-hot sampling strategy, opens up new research directions for fast acquisition of other data modalities in computer graphics, e.g. BRDFs, BTF, and light field videos. Multidimensional sensing and dictionary learning allows us to have different sampling strategies for each dimension. For instance, one can choose to take full spectral information and sub-sample the angular information. Hence, our $n$D formulation facilitates the design of new visual data acquisition devices. Another interesting aspect of the proposed method is the utilization of learned dictionaries, which we show to outperform analytical dictionaries such as DCT. 

\section*{\uppercase{Acknowledgements}}

This project has received funding from the European Union’s Hori-
zon 2020 research and innovation program under Marie Skłodowska-
Curie grant agreement No956585. We thank the anonymous reviewers for their feedback.

\bibliographystyle{apalike}
{\small
\bibliography{main}}

\end{document}